\pdfoutput=1

\documentclass[11pt]{article}

\usepackage[final]{acl}

\usepackage{times}
\usepackage{graphicx}
\usepackage{latexsym}

\usepackage[T1]{fontenc}

\usepackage[utf8]{inputenc}

\usepackage{microtype}

\newcommand{\name}{\texttt{DictaLM}}

\newcommand{\hiding}[1]{}

\title{Introducing {\name} - A Large Generative Language Model for Modern Hebrew}

\author{Shaltiel Shmidman\textsuperscript{1,†}, Avi Shmidman\textsuperscript{1,2,‡}, Amir David Nissan Cohen\textsuperscript{2,†}, Moshe Koppel\textsuperscript{1,2,†} \\
\textsuperscript{1}DICTA / Jerusalem, Israel \quad
\textsuperscript{2}Bar Ilan University / Ramat Gan, Israel \\ 
\texttt{\small \textsuperscript{†}\{shaltieltzion,moishk,amirdnc\}@gmail.com} \\
\texttt{\small \textsuperscript{‡}avi.shmidman@biu.ac.il}}

\begin{document}
\maketitle
\begin{abstract}

We present {\name}, a large-scale language model tailored for Modern Hebrew. Boasting 7B parameters, this model is predominantly trained on Hebrew-centric data. As a commitment to promoting research and development in the Hebrew language, we release both the foundation model and the instruct-tuned model under a Creative Commons license\footnote{For specifics on the license, visit \url{https://creativecommons.org/licenses/by-sa/4.0/}}. Concurrently, we introduce \name\texttt{-Rab}, another foundation model geared towards Rabbinic/Historical Hebrew. These foundation models serve as ideal starting points for fine-tuning various Hebrew-specific tasks, such as instruction, Q\&A \cite{Cohen2023HeQ}, sentiment analysis \cite{amram-etal-2018-representations}, and more \cite{barekettsarfatinemo}. This release represents a preliminary step, offering an initial Hebrew LLM model for the Hebrew NLP community to experiment with.
\end{abstract}

\section{Introduction}
Language models have revolutionized the realm of natural language processing, facilitating significant advancements in tasks ranging from sentiment analysis to machine translation. As the breadth and depth of these models expand, so does the aspiration for linguistic diversity. Yet, while the majority of state-of-the-art models cater predominantly to widely spoken languages, there exists a vast landscape of languages and dialects that are underrepresented in currently existing  large-scale language models. Hebrew is one such language.

\begin{figure}[hbt!] 
\centering
\includegraphics[scale=0.35]{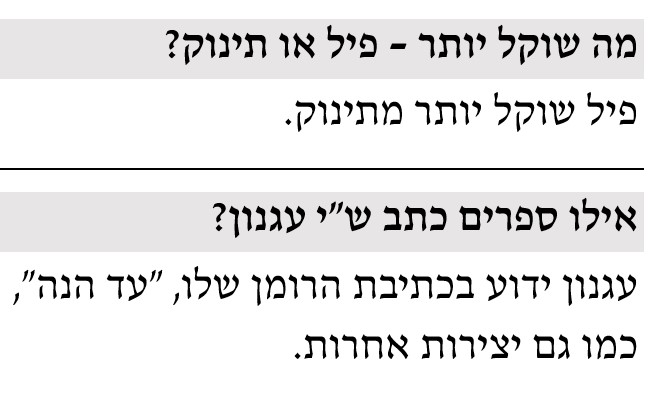} \caption{We present two instances of {\name} utilization: in the first instance, the model exhibits common sense reasoning, while in the second, it displays worldly knowledge.} \label{fig:example}
\end{figure}

In this paper, we make strides to bridge this gap by introducing {\name} - the first large-scale language model crafted for Modern Hebrew. By leveraging a dataset dominated by Hebrew-centric content, our endeavor was not only to construct a model adept at understanding and generating Modern Hebrew but also to lay down a foundation that facilitates further advancements in the field. As part of this initiative, we also present \name\texttt{-Rab}, a parallel model pretrained for Rabbinic/Historical Hebrew, thereby encompassing the vast chronological spectrum of the Hebrew language.
This release serves as a preliminary step, providing an initial tentative version to the Hebrew NLP community as a foundation for further refinements, adaptations, and collaborative enhancements. Figure \ref{fig:example} demonstrates example output from the instruct-tuned model.

\section{Datasets}
In this section, we elucidate the datasets employed for training and fine-tuning {\name}. The assemblage of data, amassing a total of 7.5 billion tokens, originates from a mixture of authentic sources; no synthetic data was added. The pre-training phase is followed by a fine-tuning stage through instruct datasets derived from Hebrew Question-Answering datasets and a translated version of the MPT Instruct Dataset. 
\subsection{Pre-training Data}
The dataset is built up of several different components:

\textbf{C4 [80\%]}. We start with the HeDC4 corpus released by \cite{shalumov2023hero}, and continue further cleaning it. We removed approximately 15\% of the corpus using various techniques including histograms, gibberish detectors, as well as removing sentences that had a very high perplexity when running through a Modern Hebrew BERT model. In addition, we limited our training corpus to contain only words in English and Hebrew, and all other languages were reduced to a designated \textit{<foreign>} token to avoid cluttering the tokenizer with non-Hebrew tokens. The resulting corpus contains approximately 6B byte-pair tokens. 

\textbf{Other sources [20\%]}. We collected data from various other sources including news sites, blogs, tv and movie subtitles, novels, and more. This data was also run through a similar cleaning process to the C4 corpus, as described above, and resulted in an additional 1.5B byte-pair tokens.

\subsubsection{Instruct Data}

Our instruct-tuning data contains a mixture of 2 different datasets, each processed and modified in order to teach the model to follow as many different instructions as possible.

\textbf{QA Datasets}. We take the HeQ \cite{Cohen2023HeQ} and ParaShoot \cite{keren2021parashoot} training datasets and format them as instructions. The prompt contains the context paragraph followed by the question, with a system instruction. The system instruction starts with a general instruction (in Hebrew) stating "Please read the following paragraph and answer the question that comes after", and 60\% of the time also instructs the system to format a specific type of response (e.g., "Short and to the point", "Please cite the sentence to support your answer", and more). We list a few examples in Appendix \ref{sec:appendix_a}.

\textbf{Translated MPT Instruct}. We took the MPT Instruct Dataset from huggingface\footnote{\url{https://huggingface.co/datasets/mosaicml/dolly_hhrlhf}} and ran it through a translation API. We then reformatted the prompt to remove the constant structure, and left the question only. We then added in each question three times: Once with no system prompt, and twice with two different prompts chosen based on the length of the response, asking the model to be concise, expand, answer in X sentences, etc. We list a few examples in Appendix \ref{sec:appendix_b}.

\section{Model architecture}
\subsection{Tokenizer}

A major problem we encountered when attempting to use other multilingual LLMs for Hebrew was the tokenization. When the corpus contains a very small percentage of a language, then the number of tokens representing that language in the vocabulary is significantly reduced. In addition, due to the nature of UTF-8 encoding, byte-pair tokenization methods result in even scarcer representation of Hebrew in the vocabulary. As can be seen in OpenAI's GPT-3 tokenizer\footnote{\url{https://platform.openai.com/tokenizer}}, if one inserts a few paragraphs of Hebrew text, the tokenizer will average 1.1 tokens per \textbf{character}. 

We train our tokenizer using the byte-pair encoding (BPE) algorithm \cite{DBLP:journals/corr/SennrichHB15} on our cleaned corpus with a vocabulary size of 56000. The resulting tokenizer had a ratio of approximately 1.3 tokens per \textbf{word}.

\subsection{Architecture}
In this section, we detail the architectural framework of {\name}. Following recent work on large language models, our network is based on the transformer architecture \cite{DBLP:journals/corr/VaswaniSPUJGKP17}. Our architecture encompasses several enhancements aimed at boosting training stability and overall performance:

\textbf{Normalization}. To improve training stability and balance the input, we normalize the input of each transformer layer before and after the attention calculation. We use the LayerNorm1P normalization with $\epsilon=1e-5$, which is a slightly modified version of the \textit{FastLayerNorm} normalization offered by NVIDIA's APEX library\footnote{\url{https://github.com/NVIDIA/apex}}.

\textbf{GeLU Activation}. As reported by \cite{hendrycks2023gaussian}, we use the GeLU activation function.\footnote{We considered using other activations (such as SwiGLU \cite{shazeer2020glu}), but in the end we went with GeLU}

\textbf{Rotary Embeddings}. Shown to be effective for extending the sequence length without a performance trase-off, we use rotary positional embedding (RoPE) with a $0.5\%$ dimension percentage, introduced by \cite{su2022roformer}, at each layer of the network. 

\textbf{Separate embedding and output weights}. As shown by \cite{welch2020improving}, separating the embeddings and the output weights leads to better performance. 

\subsection{Training Details and Hyperparameters}

We trained our model using the NeMo framework\footnote{\url{https://github.com/NVIDIA/NeMo}} which is highly optimized for training compute-heavy machine learning models on NVIDIA hardware. We pre-trained the model on 8 H100 GPUs with tensor parallel size of 2 for a total of 150 hours completing 2.5 epochs ($\sim$18.5B tokens), and then fine-tuning for instructions for 8 hours. The training was done in a combination of bf16 and fp8 precision using NVIDIA's transformer engine\footnote{\url{https://github.com/NVIDIA/TransformerEngine}}. 
The training was done with a global batch size of 128. We used the FusedAdam optimizer, with an initial learning rate of $0.00016$, betas of $0.9,0.95$ and the Cosine-Annealing schedule with a warmup of 750 steps and a minimum learning rate of $1e-5$. The details for the model size are listed in Table \ref{tab:table-hp}.

\begin{table}[h]
\centering
\begin{tabular}{|c|c|}
\hline
\textbf{Max Sequence Length}  & 2048  \\ \hline
\textbf{Num Layers}           & 32    \\ \hline
\textbf{Hidden Size}          & 4096  \\ \hline
\textbf{Intermediate Size}    & 10880 \\ \hline
\textbf{Attention Heads}      & 32    \\ \hline
\end{tabular}
\caption{Model size}
\label{tab:table-hp}
\end{table}

\subsection{\name\texttt{-Rab} Model}

In addition to the model we described above, we also trained a model \name\texttt{-Rab} for use with Rabbinic Hebrew tasks. We used the same approach as above, adjusting the input corpus to contain a large sampling of Rabbinic Hebrew data.

Specifically, we added a corpus of 1.2B tokens of Rabbinic Hebrew texts taken from various sources (e.g. Sefaria\footnote{\url{https://www.sefaria.org.il/}}, Dicta\footnote{\url{https://library.dicta.org.il/}}). We combined this corpus together with the modern Hebrew corpus that we described above, sampling the data such that fifty percent of the training sequences would be from the Rabbinic Hebrew corpus (with oversampling).   

The model uses the same tokenizer as \name, and was trained for a total of 1.5 iterations ($\sim$12.5B tokens).

We are pleased to also release this foundation model, tailored to benefit researchers working on Rabbinic Hebrew. This model can be used as a base model for fine-tuning on specific tasks relevant to the Rabbinic Hebrew domain. Our internal experiments reveal encouraging results with Rabbinic texts, details of which will be shared in forthcoming publications.

\section{Drawbacks}

Our model was trained on the full dataset without any censorship for offensive or biased material, and therefore it may generate sentences that are offensive to some users. 

Also, we would like to highlight that this project is in its alpha phase. While we are releasing {\name} to facilitate research endeavors, and while we believe that it can serve as a useful foundation for specific fine-tuned tasks in the realm of Hebrew NLP, we acknowledge that the quality of the model does not yet match industry standards.

\section{Conclusion}

We are pleased to present the three models described within this paper: the two foundational models (suitable as base models for further fine-tuning for tasks concerning both Modern and Rabbinic Hebrew), and the instruct model, fine-tuned to address instruction prompts in Modern Hebrew. The public release of these models aims to contribute to the advancement of research and development within the Hebrew NLP domain. The models can be accessed via the following links:

\begin{itemize}
\item Foundation model \name: \url{https://huggingface.co/dicta-il/dictalm-7b}

\item Instruct model \name\texttt{-Instruct}: \url{https://huggingface.co/dicta-il/dictalm-7b-instruct}

\item Foundation model for Rabbinic Hebrew \name\texttt{-Rab}: \url{https://huggingface.co/dicta-il/dictalm-rab-7b}

\end{itemize}

\bibliography{anthology,custom}
\bibliographystyle{acl_natbib}

\newpage
\onecolumn

\appendix

\section{Appendix: Instruct Examples from QA Datasets}
\label{sec:appendix_a}

\begin{figure}[hbt!] 
\includegraphics[page=1, scale=1]{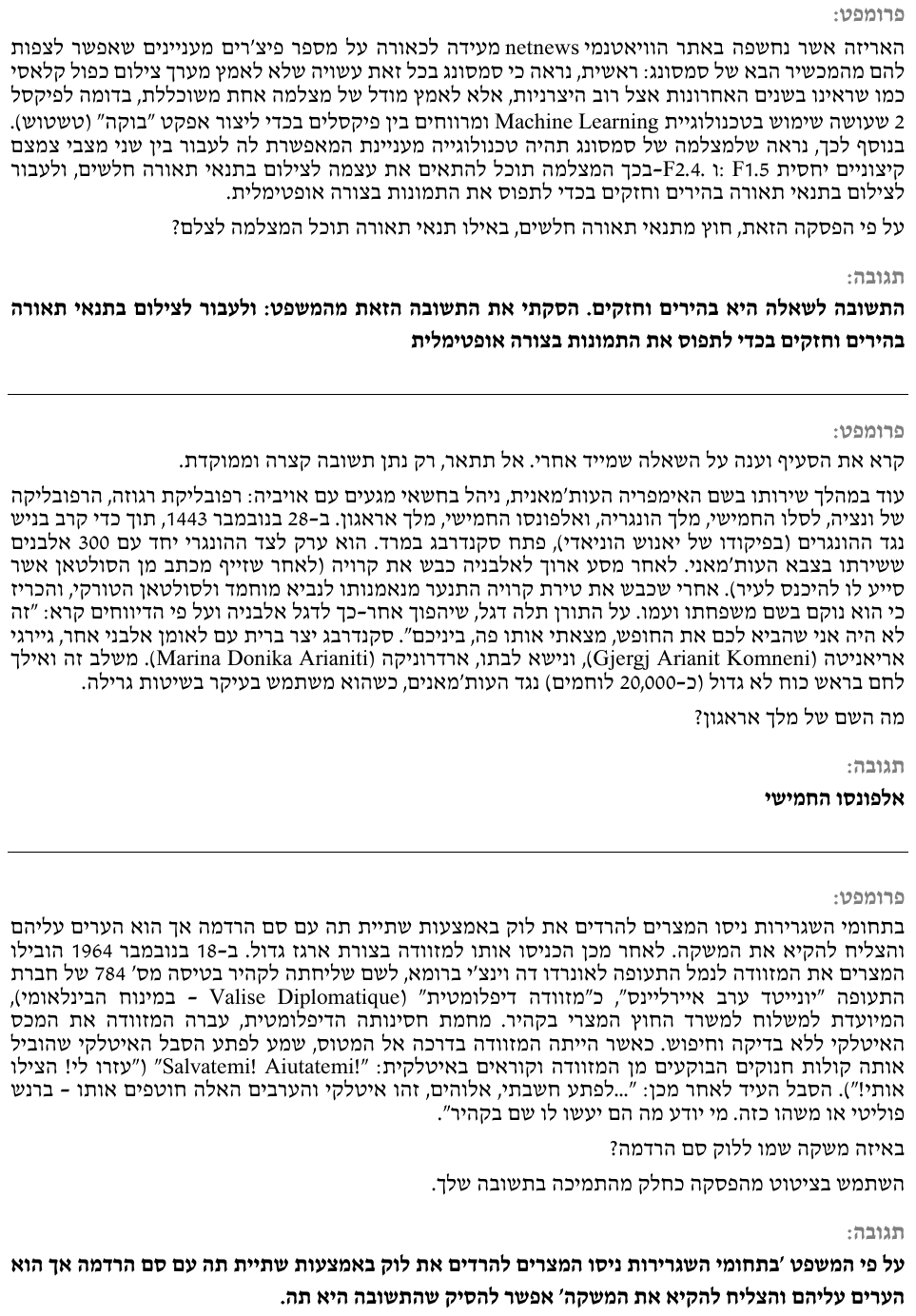}
\end{figure}

\section{Appendix: Instruct Examples from Translated MPT-Instruct}
\label{sec:appendix_b}

\begin{figure}[hbt!] 
\includegraphics[page=2, scale=1]{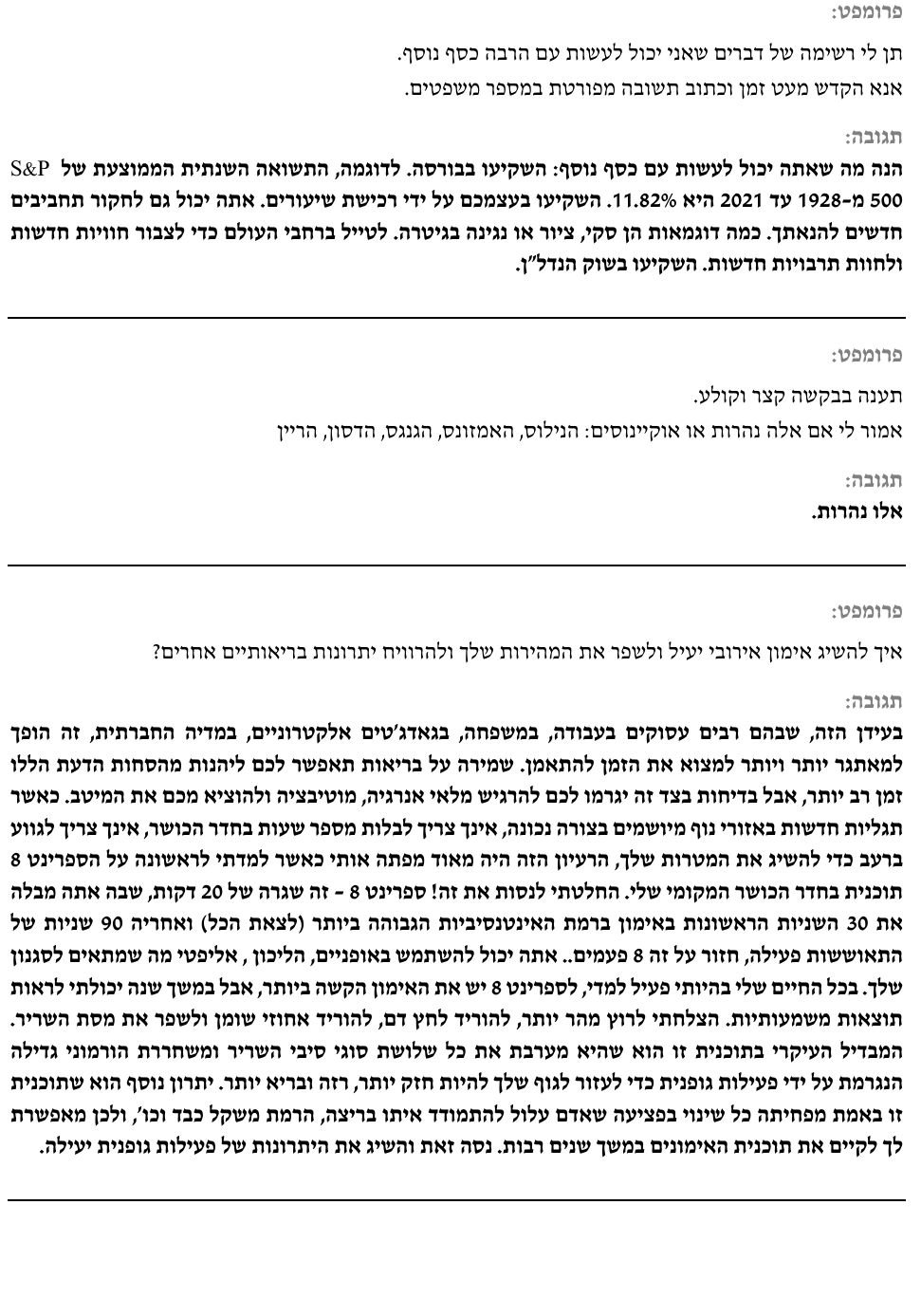}
\end{figure}

\end{document}